\documentclass[runningheads]{llncs}
\usepackage{graphicx}
\usepackage{comment}
\usepackage{amsmath,amssymb} 
\usepackage{color}

\usepackage{subfig}
\usepackage{booktabs}
\usepackage{multirow}
\usepackage{dblfloatfix}    
\usepackage{array}
\usepackage[font=small,labelfont=bf]{caption}

\newcommand{\drule}{\specialrule{0.2pt}{1pt}{1pt}%
            \specialrule{0.2pt}{0pt}{\belowrulesep}%
            }
\newenvironment{myindentpar}[1]%
  {\begin{list}{}%
          {\setlength{\leftmargin}{#1}}%
          \item[]%
  }
  {\end{list}}
\usepackage{kotex}

\newcommand\blfootnote[1]{%
  \begingroup
  \renewcommand\thefootnote{}\footnote{#1}%
  \addtocounter{footnote}{-1}%
  \endgroup
}

\begin{document}
\pagestyle{headings}
\mainmatter
\def\ECCVSubNumber{100}  

\title{Towards Lightweight Lane Detection by Optimizing Spatial Embedding} 

\titlerunning{Towards Lightweight Lane Detection}
%
\author{Seokwoo Jung\inst{\text{*}1} \and
Sungha Choi\inst{\text{*}1} \and
Mohammad A. Khan\inst{2} \and
Jaegul Choo\inst{2}}
\authorrunning{S. Jung et al.}
%
\institute{A\&B Center, LG Electronics,\, Seoul, South Korea \\
\email{\{seokwoo.jung, sungha.choi\}@lge.com}\\
\and
KAIST,\, Daejeon, South Korea\\
\email{muhammedazamkhan@gmail.com, jchoo@kaist.ac.kr}}
\maketitle
\blfootnote{\vspace*{-0.3cm}* indicates equal contribution}

\begin{abstract}
\vspace*{-1.0cm}
A number of lane detection methods depend on a proposal-free instance segmentation because of its adaptability to flexible object shape, occlusion, and real-time application. This paper addresses the problem that pixel embedding in proposal-free instance segmentation based lane detection is difficult to optimize. A translation invariance of convolution, which is one of the supposed strengths, causes challenges in optimizing pixel embedding. In this work, we propose a lane detection method based on proposal-free instance segmentation, directly optimizing spatial embedding of pixels using image coordinate. Our proposed method allows the post-processing step for center localization and optimizes clustering in an end-to-end manner. The proposed method enables real-time lane detection through the simplicity of post-processing and the adoption of a lightweight backbone. Our proposed method demonstrates competitive performance on public lane detection datasets.
\vspace*{-0.3cm}
\end{abstract}

\vspace*{-0.8cm}
\section{Introduction}
\vspace*{-0.3cm}
Lane marker detection, a fundamental task in autonomous driving, has accomplished substantial advancements through the adoption of convolutional neural networks (CNNs). A few studies formulate the lane marker detection task into semantic segmentation~\cite{chen2018encoder} or instance segmentation~\cite{he2017mask,neven2019instance} problem in terms of segmenting the lane area and classifying the lane to which each pixel belongs. 
Several studies have recently suggested lane detection methods~\cite{hsu2018learning,neven2018towards} based on proposal-free instance segmentation~\cite{neven2019instance,de2017semantic} and demonstrated remarkable performance.
Although the current dominant methods for instance segmentation are proposal-based (i.e., detect-then-segment approach)~\cite{he2017mask},
proposal-free methods are more suitable for lane detection. The proposal-free approaches 
map pixels into an embedding space utilizing CNNs, and learn embedding to close points if the pixels belong to the same instance. 
These approaches have strengths in long, thin shapes and complex occlusion, which are not suitable for bounding box, and real-time application.
Additionally, several proposal-free approaches~\cite{de2017semantic,neven2018towards} utilize a discriminative loss in the form of triplet loss~\cite{Schroff_2015_CVPR} that pulls samples belonging to the same instance closer and pushes clusters away from each other to facilitate the learning of the pixel embedding.

The proposal-free method can be considered as a coordinate transform problem~\cite{liu2018intriguing}, which requires learning a mapping between coordinates in image pixel and another embedding space. However, CoordConv~\cite{liu2018intriguing} shows that CNNs fail to transform spatial representations between two different types of space. This phenomenon is caused by the translation-invariant nature of convolution, which in turn makes pixel embedding difficult to optimize. Recent studies~\cite{liu2018intriguing,neven2019instance} alleviate the complexity in learning the pixel embedding by exploiting coordinate information of input.

Inspired by these work~\cite{liu2018intriguing,neven2019instance,de2017semantic,neven2018towards}, we propose a novel lane marker detection method based on instance segmentation approach with making cluster-friendly spatial embedding. The contributions of this paper include:

\begin{myindentpar}{0.2cm}
\vspace*{-0.1cm}
\noindent$\bullet$ \,To the best of our knowledge, this is the first endeavor that adopts spatial embedding of pixels using image coordinate that overcomes the limitation of translation invariant nature of convolution to solve lane detection task.
\vspace*{-0.1cm}
\\[0.3em]
$\bullet$ \,Our proposed method enables real-time lane detection through 
immediately predicting the center and elements of lane instance with fast iterative process and adopting a lightweight backbone. Nonetheless, our method shows competitive performance compared to other state-of-the-art methods.
\end{myindentpar}

\vspace*{-0.5cm}
\section{Method}
\vspace*{-0.2cm}
This section describes the architecture of our proposed method, loss function and post-processing. The proposed lane detection method is based on the state-of-the-art proposal-free instance segmentation~\cite{neven2019instance}.

\vspace*{-0.55cm}
\begin{figure*}[ht!]
  \centering\includegraphics[width=\linewidth]{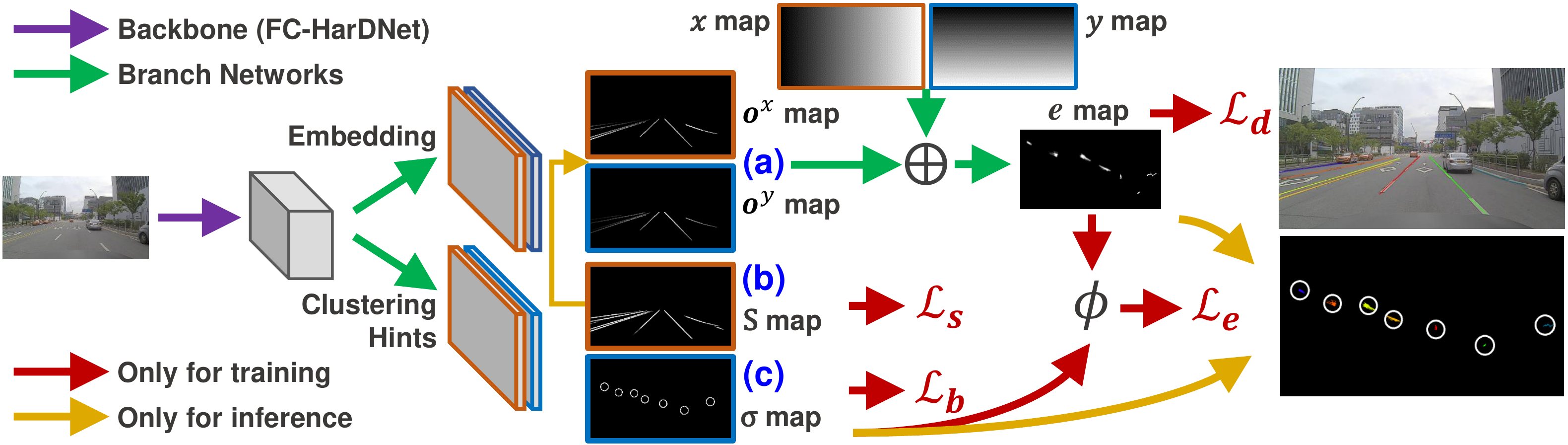}
\vspace*{-0.6cm}
  \caption{Overall pipeline of our model. The details are described in Section~\ref{model_overview}.
}
\label{fig:overall_architecture}
\vspace*{-0.9cm}
\end{figure*}
\vspace*{-0.05cm}
\subsection{Model overview}\label{model_overview}
\vspace*{-0.05cm}
We adpopt FCHarDNet~\cite{Chao_2019_ICCV}, a state-of-the-art in terms of computational efficiency and low memory-traffic, as backbone networks for our proposed model. On top of the backbone, we attach two separate branches dedicated to predict spatial embedding and clustering hints.
The embedding branch predicts the offset ($o^x$ and $o^y$ map) for the pixels in an image, pointing to the center of the lane instance to which the pixel belongs. The spatial embedding vectors ($e$ map) for the pixels are defined as $e=[x+o^{x};y+o^{y}]$, where $x$ and $y$ map denote the coordinates of the pixels. So the spatial embedding $e_i$ for each pixel $i$ is assigned to the centroid of the lane instance to which the pixel belongs.
The clustering hints branch infers two outputs: the first is probability map ($S$ map) for the center of lane instance and the second is the clustering bandwidth ($\sigma$ map) related to the size of each lane instance.

\vspace*{-0.3cm}
\subsection{Loss}
\vspace*{-0.16cm}
We adopt Lovasz-hinge loss~\cite{yu2015lovsz} and a Gaussian function to learn the offset (Fig.~\ref{fig:overall_architecture}(a)) and clustering bandwidth (Fig.~\ref{fig:overall_architecture}(c)) as proposed by Neven et al. (2019)~\cite{neven2019instance}. The Gaussian function is defined as $\phi_k(e_i)=\text{exp}(-\frac{\|e_i-C_k\|^2}{2\sigma_k^{2}})$,
where 
$C_k$ denotes the centroid of $k$-th instance, and $\sigma_k$ is the average of all $\sigma_i$ belonging to k-th instance. $\phi_k(e_i)$ derives the probability that $e_i$ belongs to the $k$-th instance from the $\sigma_k$ and the distance between $e_i$ and $C_k$. If only $e_i$ with $\phi_k(e_i)\geq \Pr$ are assigned to $k$-th instance, the margin is calculated as $\sqrt{-2\sigma_k^{2}\ \ln\Pr}$, where $\Pr$ denotes the probability threshold. The embedding loss function $\mathcal{L}_e$ is defined as

\vspace*{-0.65cm}
\begin{equation} \label{eq:Loss_e}
\nonumber
\mathcal{L}_e = \frac{1}{K}\sum_{k=1}^{K}\mathcal{L}_h(\left\{y \right\}, \left\{ \phi_k(e_i) \right\}), \quad
y =\begin{cases}
 1, & \text{if $e_i \in S_k$}\\
 0, & \text{otherwise}
\end{cases} \quad
\forall e_i \in \text{fg}
\vspace*{-0.2cm}
\end{equation}
where $\mathcal{L}_h$ denotes Lovasz-hinge loss, and $S_k$ is $k$-th instance. Note that including background pixels for the embedding loss interferes with the spatial embedding of pixels belonging to the instance converging to the centroid. Therefore, the background pixels are not considered in the embedding loss function unlike the previous work~\cite{neven2019instance}. To increase $\phi_k(e_i)$ of pixels belonging to $k$-th instance, $\sigma_k$ grows to enlarge the margin rather than decreasing $\|e_i-C_k\|$.
To compensate for this, the bandwidth saturation loss is formulated as
\vspace*{-0.25cm}
\begin{equation} \label{eq:Loss_e}
\nonumber
\mathcal{L}_b = \frac{1}{K}\sum_{k=1}^{K} \text{max}\left( \sqrt{-2\sigma_k^{2}\ \ln\Pr} - \delta_m , \  0 \right)
\vspace*{-0.2cm}
\end{equation}

Additionally, the loss $\mathcal{L}_d$ for inter-cluster push force~\cite{de2017semantic} is adopted to minimize interference between adjacent clusters in embedding space, and the loss $\mathcal{L}_s$~\cite{neven2019instance} is applied for sampling instance center at inference time.

\begin{table}[b!]
\vspace*{-0.65cm}
\begin{center}
\footnotesize
\renewcommand{\tabcolsep}{1.0mm}
\begin{tabular}{c|c|c|c|c|c}
\toprule
\multirow{2}{*}{Method} & \multirow{2}{*}{Acc(\%)} & \multirow{2}{*}{FP} & \multirow{2}{*}{FN} & \multicolumn{2}{c}{Inference time (msec)}\\
& & & & Networks & Post-pro. \\ 
\drule
$^\dagger$Baseline (Push + Pull)~\cite{neven2018towards} & 96.42 & 0.0654 & 0.0195 & 6.62 & $^\ddagger$498\\
\midrule
$^\dagger$Baseline (SE)~\cite{neven2019instance} & 95.96 & 0.0799 & 0.0261 & N/A & N/A\\
\midrule
Ours (SE + FG) & 96.20 & 0.0598 & 0.0256 & N/A & N/A \\
\midrule
Ours (SE + FG + BS + Push) & \textbf{96.58} & \textbf{0.0540} & \textbf{0.0177} & 6.64 & \textbf{3.20} \\
\bottomrule
\end{tabular}
\end{center}
\vspace*{-0.2cm}
\caption{TuSimple benchmark results for an image size of 1280$\times$720. Push, Pull, SE, FG and BS are abbreviations for inter-cluster push, intra-cluser pull, spatial embedding, the embedding loss only for foreground pixels and bandwidth saturation. $^\dagger$ denotes our re-implementation of the baseline. $^\ddagger$ denotes adoption of DBSCAN for post-processing. The inference time is measured on NVIDIA V100 and Xeon Silver 4114.}
\label{tab_best_model}
\end{table}

\vspace*{-0.4cm}
\section{Experiments}
\begin{figure*}[ht!]
\vspace*{-0.55cm}
  \centering\includegraphics[width=0.9\linewidth]{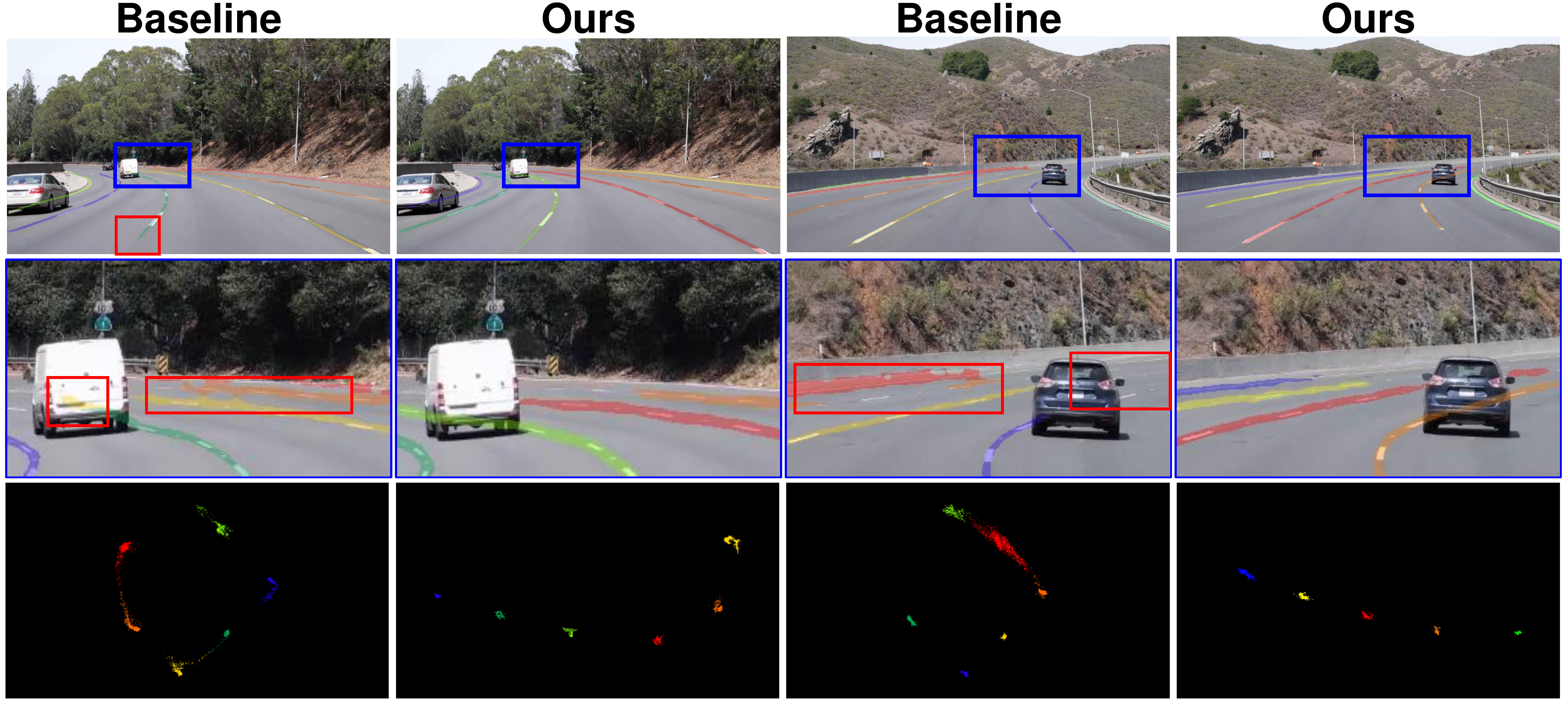}
\vspace*{-0.15cm}
  \caption{Qualitative comparison between baseline~\cite{neven2018towards} and ours on real road environment. From top to bottom: input image with predicted lane marker, magnifying the blue boxes in the image of the first row, spatial embedding of the pixels in the input images. Red boxes denote the incorrectly predicted regions.
}
\label{fig:qualitative}
\vspace*{-0.65cm}
\end{figure*}

\vspace*{-0.25cm}
In this section, we quantitatively and qualitatively compare the performance of our model with the baseline models~\cite{neven2018towards,neven2019instance} to verify the effectiveness of the proposed model. Our model is superior to the baseline models in terms of accuracy and inference speed (Table~\ref{tab_best_model}). Especially, the proposed method is much faster than the baseline due to simple post-processing. Finally, Fig.~\ref{fig:qualitative} shows the qualitative results compared to the baseline in a real road environment.


\vspace*{-0.42cm}
\section{Conclusions}
\vspace*{-0.18cm}
We proposed novel methods for applying spatial embedding to lane detection, and experimentally verified that our methods is efficient compared to state-of-the-arts models. Furthermore, we will enhance our methods through further extensive experiments and for real-time applications in an embedded device.
\vspace*{-0.28cm}
%
%
{\footnotesize
\bibliographystyle{splncs04}
\bibliography{egbib}

\begin{thebibliography}{10}
\providecommand{\url}[1]{\texttt{#1}}
\providecommand{\urlprefix}{URL }
\providecommand{\doi}[1]{https://doi.org/#1}

\bibitem{Chao_2019_ICCV}
Chao, P., Kao, C.Y., Ruan, Y.S., Huang, C.H., Lin, Y.L.: Hardnet: A low memory
  traffic network. In: ICCV (October 2019)

\bibitem{chen2018encoder}
Chen, L.C., Zhu, Y., Papandreou, G., Schroff, F., Adam, H.: Encoder-decoder
  with atrous separable convolution for semantic image segmentation. In: ECCV
  (2018)

\bibitem{de2017semantic}
De~Brabandere, B., Neven, D., Van~Gool, L.: Semantic instance segmentation with
  a discriminative loss function. arXiv preprint arXiv:1708.02551  (2017)

\bibitem{he2017mask}
He, K., Gkioxari, G., Doll{\'a}r, P., Girshick, R.: Mask r-cnn. In: ICCV (2017)

\bibitem{hsu2018learning}
Hsu, Y.C., Xu, Z., Kira, Z., Huang, J.: Learning to cluster for proposal-free
  instance segmentation. In: IJCNN. IEEE (2018)

\bibitem{liu2018intriguing}
Liu, R., Lehman, J., Molino, P., Such, F.P., Frank, E., Sergeev, A., Yosinski,
  J.: An intriguing failing of convolutional neural networks and the coordconv
  solution. In: Advances in Neural Information Processing Systems. pp.
  9605--9616 (2018)

\bibitem{neven2019instance}
Neven, D., Brabandere, B.D., Proesmans, M., Gool, L.V.: Instance segmentation
  by jointly optimizing spatial embeddings and clustering bandwidth. In: CVPR
  (2019)

\bibitem{neven2018towards}
Neven, D., De~Brabandere, B., Georgoulis, S., Proesmans, M., Van~Gool, L.:
  Towards end-to-end lane detection: an instance segmentation approach. In:
  2018 IEEE intelligent vehicles symposium (IV). pp. 286--291. IEEE (2018)

\bibitem{Schroff_2015_CVPR}
Schroff, F., Kalenichenko, D., Philbin, J.: Facenet: A unified embedding for
  face recognition and clustering. In: CVPR (June 2015)

\bibitem{yu2015lovsz}
Yu, J., Blaschko, M.: The lovász hinge: A novel convex surrogate for
  submodular losses (2015)

\end{thebibliography}
}
\end{document}